\documentclass[final,5p,times,twocolumn]{elsarticle}

\usepackage{amsmath}
\usepackage{enumitem}
\usepackage{pdflscape}
\usepackage{color}
\definecolor{chred}{rgb}{0.8,0,0}
\usepackage{graphicx}
\usepackage{dblfloatfix}
\usepackage{subfigure}
\usepackage{eqlist}
\usepackage{txfonts}
\usepackage{url}
\usepackage[lined, ruled, linesnumbered]{algorithm2e}
\usepackage{footmisc}
\usepackage{booktabs}

\journal{Neurocomputing}

\begin{document}

\begin{frontmatter}



\title{Teaching Robots to Do Object Assembly using Multi-modal 3D
Vision}


\author[label1]{Weiwei Wan}
\author[label2]{Feng Lu}
\author[label1]{Zepei Wu}
\author[label1]{Kensuke Harada}
\address[label1]{National Institute of Advanced Industrial Science and
Engineering, Japan}
\address[label2]{Beihang University, China}

\begin{abstract}
The motivation of this paper is to develop a smart system using multi-modal
vision for next-generation mechanical assembly. It includes two phases where in
the first phase human beings teach the assembly structure to a robot and in the
second phase the robot finds objects and grasps and assembles them using AI
planning. The crucial part of the system is the precision of 3D visual detection
and the paper presents multi-modal approaches to meet the requirements: AR
markers are used in the teaching phase since human beings can actively control
the process. Point cloud matching and geometric constraints are used in the
robot execution phase to avoid unexpected noises. Experiments are performed to
examine the precision and correctness of the approaches. The study is
practical: The developed approaches are integrated with graph
model-based motion planning, implemented on an industrial
robots and applicable to real-world scenarios.

\end{abstract}


\end{frontmatter}


\section{Introduction}
\label{intro}
The motivation of this paper is to develop a smart system using multi-modal
vision for next-generation mechanical assembly:
A human worker assembles mechanical parts in front of a vision system;
The system detects the position and orientation of the assembled parts and
learns how to do assembly following the human workerfs demonstration; An
industrial robot performs assembly tasks following the data learned from human
demonstration; It finds the mechanical parts in its workspace, picks up them,
and does assembly using motion planning and assembly planning.

The difficulty in developing the smart system is precise visual detection. Two
problems exist where the first one is in the human teaching phase, namely how to
precisely detect the position and orientation of the parts in human hands
during manual demonstration; The second one is in the robot execution phase,
namely how to precisely detect the position and orientation of the parts in
the workspace and perform assembly.

Lots of detecting and tracking studies are available in contemporary
publication, but none of them meets the requirements of the two problems. The
approaches used include RGB images, markers, point cloud,
extrinsic constraints, and multi-modal solutions where the RGB images and
markers are short in occlusions, the point cloud and extrinsic constraints are
short in partial loss and noises, and the multi-modal solutions are not clearly
stated and are still being developed.

This paper solves the two problems using multi-modal vision. 
First, we attach AR markers to the objects for
assembly and track them by detecting and transforming the marker positions
during human demonstration. We don't need to worry occlusions since the teaching
phase is manual and is performed by human beings who are intelligent enough to
actively avoid occlusions and show good features to vision systems. The modal
employed in this phase is the markers (rgb image) and the geometric relation
between the markers and the object models. The tag ``AR(RGB)'' is used for
representation.

Second, we use depth cameras and match the object model to the point cloud
obtained using depth camera to roughly estimate the object pose, and use the
geometric constraints from planar table surface to reinforce the precision.
For one thing, the robot execution phase is automatic and is not as flexible as
human teaching. Markerless approaches are used to avoid occlusions. For
the other, the point cloud and extrinsic constraints are fused to make up the
partial loss and noises. The assumption is when the object is placed on the
surface of a table, it stabilizes at a limited number of poses inherent to the
geometric constraints. The poses help to freeze some Degree of Freedom and
improve precision. The modal employed in this phase is the cloud point data
and the geometric constraints from the plenary surface. The tag ``Depth+Geom''
is used for representation.

Moreover, we propose an improved graph model based on our previous work to
perform assembly planning and motion planning.
Experiments are performed to examine the precision and correctness of our
approaches. We quantitatively show the advantages of ``AR(RGB)'' and
``Depth+Geom'' in next-generation mechanical assembly and concretely demonstrate
the process of searching and planning using the improve graph model. We also
integrate the developed approaches with Kawada Nextage Robots and
show the applicability of them in real-world scenarios.

\section{Related Work}
\label{relwork}

This paper is highly related to
studies in 3D object detection for robotic manipulation and assembly and the
literature review concentrates on the perception aspect. For general studies
on robotic grasping, manipulation, and assembly, refer to \cite{Handeybook},
\cite{Mason01}, and \cite{Dogar15}.
The literature review emphasizes on model-based approaches since the paper is
motivated by next-generation mechanical assembly and is about the industrial
applications where precision is crucial and object models are available. For
model-less studies, refer to \cite{Goldfeder10} and \cite{Lenz14}. For
appearance-based studies, refer to \cite{Murase95} \cite{Mittrapiyanuruk04}, and
\cite{Zickler06}.
Moreover, learning approaches are no reviewed since they are not precise.
Refer to \cite{Stark10} and \cite{Libelt10} if interested.

We archive and review the related work
according to the modals used for 3D detection, including RGB images,  markers,
point cloud, haptic sensing, extrinsic constraints, and multi-modal fusion.

\subsection{RGB images}
RGB images are the most commonly used modal of robotic perception.
Using RGB images to solve the model-based 3D position and orientation detection
problem is widely known as the ``model-to-image registration problem''
\cite{Wunsch96} and is under the framework of POSIT (Pose from Orthography and
Scaling with Iterations) \cite{David02}. When looking for objects in images,
the POSIT-based approaches match the feature descriptors by comparing the most
representative features of an image to the features of the object for
detection. The features could either be values computed at pixel points or
histograms computed on a region of pixels. Using more than three matches, the 3D
position and orientation of an object can be found by solving polynomial
equations \cite{Fischler81, DeMenthon95, Lu00}. A good material that explains
the matching process can be found in \cite{Schaffalitzky02}. The paper studies
multi-view image matching. It is not directly related to 3D detection, but
explains well how to match the feature descriptors.

Some of the most common choices of features include corner features
\cite{Harris88} applied in \cite{Chia02} and \cite{David02}, line features
applied in \cite{David03}, \cite{Klein03} and \cite{Marchand02}, cylinder
features applied in \cite{Marchand02} and \cite{Harada13}, and SIFT features
\cite{Lowe04} applied in \cite{Gordon06} and \cite{Collet09}. Especially,
\cite{Gordon06} stated clearly the two stages of model-based detection using
RGB images: (1) The modeling stage where the textured 3D model is recovered from
a sequence of images; (2) The detection stage where features are extracted and
matched against those of the 3D models. The modeling stage is based on the
algorithms in \cite{Schaffalitzky02}. The detection stage is open to different
features, different polynomial solving algorithms, and some optimizations like
Levenberg-Marquardt \cite{Press92} and Mean-shift \cite{Cheng95}, etc.
\cite{Ramisa12} compared the different algorithms in the second stage.

\subsection{Markers}
In cases where the objects don't have enough features, markers are used
to assist image-based detection. Possible marker types include: AR markers,
colored markers, and shape markers, etc. The well known AR libraries \cite{Fiala05,
Wagner07} provide robust libraries and the Optitrack device provides easy-to-use
systems for the different marker types. However, the applications with markers
require some manual work and there are limitations on marker sizes, view
directions, etc.

\cite{Sundareswaran98} uses circular concentric ring fiducial markers which are
placed at known locations on a computer case to overlay the hidden innards of
the case on the camerafs video feed. \cite{Kato99} uses AR markers to locate the
display lots during virtual conference. It explains the underneath computation
well. \cite{Vacchetti04} doesnft directly use markers, but uses the offline
matched keyframes, which are essential the same thing, to correct online
tracking. \cite{Makita12} uses AR markers to recognize and cage objects.
\cite{Suligoj13} uses both balls (circles) and AR markers to estimate and track
the position of objects. More recently, \cite{Ramirezamaro15} uses AR markers to
track the position of human hands and the operating tools and uses the tracked
motion to teach robots. The paper shares the same assumption that human
beings can actively avoid occlusions. However, it doesn't need and didn't
analyze the precision since the goal of tracking is in task level, instead of
the low level trajectories.

\subsection{Point cloud}
Point cloud can be acquired using a structured light camera \cite{Freedman12},
stereovision camera \cite{Choi12}, and LIDAR device \cite{Bauhahn09}, etc. 
The recent availability of low cost depth sensors and the handy Point
Cloud Library (PCL) \cite{Rusu11} has widely disseminated 3D sensing technology in
research and commercial applications. The basic flow of using point cloud for
detection is the same as image-based detection: The first step is to extract
features from models and objects; The second step is to match the features and
detect the 3D pose.

\cite{Shutz97} is one of the early studies that uses point cloud to detect the
pose of an object. It is based on the Iterative Closest Point (ICP) algorithm
\cite{Besl92} which iteratively minimizes the mean square distance of nearest
neighbor points between the model and the point cloud. Following work basically
uses the same technique, with improvements in feature extraction and matching
algorithms. The features used in point cloud detection are more vast than those
in image-based detection, including local descriptors like signature of
histograms of orientation (SHOT) \cite{Tombari10} and Radius-based Surface
Descriptor (RSD) \cite{Marton11}, and global descriptors like Clustered
Viewpoint Feature Histogram (CVFH) \cite{Aldoma11} and Ensemble of Shape
Functions (ESF) \cite{Wohlkinger11}. The matching algorithms didnft change much,
sticking to Random Sample Consensus (RANSAC) \cite{Fischler81} and ICP. A
complete review and usage of the features and matching algorithms can be found
in \cite{Aldoma12}.

\subsection{Extrinsic constraints}

Like the markers, extrinsic constraints are used to assist point cloud based
detection. When the point cloud don't provide enough features or when the
point cloud are noisy and occluded, it is helpful to take into account the
extrinsic constraints. For example, the detection pipeline in \cite{Aldoma12}
uses hypothesis, which is one example of extrinsic constraints, to verify the
result of feature matching. \cite{Shiraki14} analyzes the functions of connected
object parts and uses them to refine grasps. The paper is not directly related
to detection but is an example of improving performance using extrinsic
constraints caused by adjacent functional units.

Some other work uses geometric constraints to reduce ambiguity of ICP.
\cite{Schuster10} and \cite{Somanath09} segment 3D clutter using the geometric
constraints. They are not directly related to detection but are used widely as
the initial steps of many detection algorithms. \cite{Savalcalvo15} clusters
point cloud into polygon patches and uses RANSAC with the multiple polygon
constraints to improve the precision of detection. \cite{Goron12} uses table
constraints for segmentation and uses Hough voting to detect object poses.
\cite{Cheung15} uses the sliced 2D contours of 3D stable placements to reduce
the noises of estimation. It is like our approach but is contour-based and
suffers from ambiguity.

\subsection{Multi-modal fusion}

Multi-modal approaches are mostly the combination or repetition of the previous
five modals. For example, some work fuses repeated modals to improve object
detection. \cite{Taylor03} fuses colour, edge and texture cues predicted from
a textured CAD model of the tracked object to recover the 3D pose, and is open
to additional cues.

Some other work uses visual tracking to correct the noises caused by fast
motions and improve the precision of initial matches. The fused modals include
the RGB image modal and the motion modal where the later one could be either
estimated using image sequences or third-part sensors like Global Positioning
System (GPS) or gyros. \cite{Kempter12} is one representative work which fuses
model motion (model-based tracking) and model detection in RGB images to refine
object poses. \cite{Klein03} fuses gyro data and line features of RGB images to
reinforce the pose estimation for head-mounted displays. \cite{Reitmayr06} uses
gyro data, point descriptors, and line descriptors together to improve the
performance of pose estimation for outdoor applications.

\cite{Pangercic11} uses point cloud to cluster the scene and find the
Region-of-Interests (ROIs), and uses image modal to estimate the object pose at
respective ROIs. The fused modals are RGB images and Point cloud.
\cite{Hinterstoisser11} also combines image and depth modals. It uses the image
gradients found on the contour of images and the surface normals found on the
body of point cloud to estimate object poses.

To our best knowledge, the contemporary object detection studies do not meet
our requirements about precision. The most demanding case is robotic grasping
and simple manipulation, which is far less strict than regrasp and assembly. We
develop our own approaches in this paper by fusing different modals to deal with
the problems in the teaching phase and the robot execution phase respectively.
For one thing, we use AR markers to detect the 3D object positions and
orientations during human teaching. For the other, we use the cloud point data
and the geometric constraints from planar table surface during robot execution.
The details are presented in following sections.

\section{System Overview}

We present an overview of the next-generation mechanical assembly system
and make clear the positions of the 3D detection in this section, and describe
in detail the ``AR(RGB)'' and ``Depth+Geom'' approaches in the sections
following it.

\begin{figure}[!htbp]
  \centering
  \includegraphics[width = 3.2in]{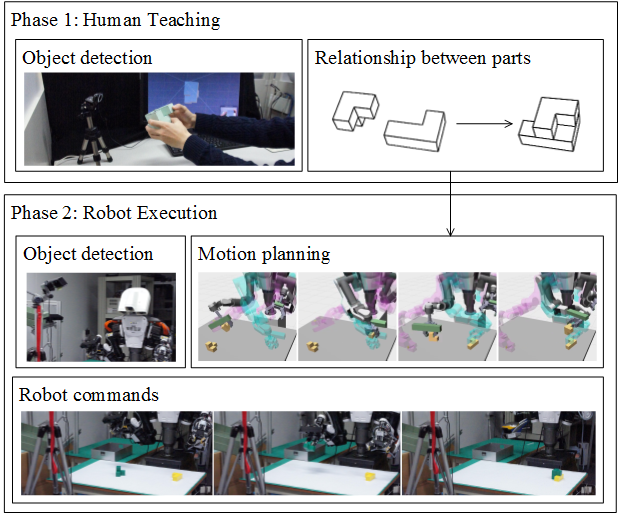}
  \caption{The flow of the next-generation mechanical assembly. The flow is
  composed of a human teaching phase and a robot execution phase. In the human
  teaching phase, a human worker demonstrates assembly with marked objects in
  front of an RGB camera. The computer detects the relationship of the assembly
  parts. In the robot execution phase, the robot detects the parts in the
  workspace using depth camera and geometric constraints, picks up them, and
  performs assembly.}
  \label{flow}
\end{figure}

Fig.\ref{flow} shows the flow of the next-generation mechanical assembly system.
It is composed of a human teaching phase and a robot execution phase. In the
human teaching phase, a human worker demonstrates how to assemble mechanical
parts in front of a vision system. The system records the relationship of the
two mechanical parts and saves it as an intermediate value.

In the robot execution phase, the robot uses another vision system to find the
mechanical parts in the workspace, picks up them, and performs assembly. The
relative position and orientation of the assembly parts are the intermediate
values perceived by the human teaching phase. The grasping configurations of
the robotic hand and the motions to move the parts are computed online using
motion planning algorithms.

The beneficial point of this system is it doesn't need direct robot programming
and is highly flexible. What the human worker needs to do is to attach the
markers and presave the pose of the marker in the object's local coordinate
system so that the vision system can compute the pose of the object from the
detected marker poses.

The 3D detection in the two phases are denoted by the two ``object detection''
sub-boxes in Fig.\ref{flow}. The one in the human teaching phase uses AR markers
since human beings can intentially avoid unexpected partial occlusions by human
hands or the other parts, and as well as ensure high precision. The one in the
motion planning phase uses point cloud data to roughly detect the object pose,
and uses the geometric constraints from planar table surface to correct the
noises and improve precision. The details will be explained in following
sections. Before that, we list the symbols to facilitate readers.

\begin{itemize}[noitemsep, nolistsep, leftmargin=0.5in]
\item[${\boldsymbol{p}_X}^s$] The position of object \textit{X} on a planery
surface. We use \textit{A} and \textit{B} to denote the two objects and
consequently use ${\boldsymbol{p}_A}^s$ and ${\boldsymbol{p}_B}^s$ to denote
their positions.
\item[${\mathbf{R}_X}^s$] The orientation of object \textit{X} on a planery
surface. Like ${\boldsymbol{p}_X}^s$, \textit{X} is to be replaced by \textit{A}
or \textit{B}.
\item[${\boldsymbol{p}_X}^a$] The position of object \textit{X} in the assembled
structure.
\item[${\mathbf{R}_X}^a$] The orientation of object \textit{X} in the assembled
structure.
\item[${\boldsymbol{p}_X}^p$] The pre-assembly positions of the two objects. The
robot will plan a motion to move the objects from the initial positions
to the preassemble positions.
\item[${\mathbf{g}_X}^f$] The force-closure grasps of object \textit{X}. The letter $f$
indicates the object is free, and is not in an assembled structure or laying on
something.
\item[${\mathbf{g}_X}^{s'}$] The force-closure grasps of object \textit{X} on a planery
surface. It is associated with ${\boldsymbol{p}_X}^s$ and ${\mathbf{R}_X}^s$.
\item[${\mathbf{g}_X}^s$] The collision-free and IK (Inverse Kinematics)
feasible grasps of object \textit{X} on a planery surface. It is also associated with
${\boldsymbol{p}_X}^s$ and ${\mathbf{R}_X}^s$.
\item[${\mathbf{g}_X}^{a'}$] The force-closure grasps of object \textit{X} in an
assembled structure. It is associated with ${\boldsymbol{p}_X}^a$ and
${\mathbf{R}_X}^a$.
\item[${\mathbf{g}_X}^a$] The collision-free and IK (Inverse Kinematics)
feasible grasps of object \textit{X} in the assembled structure. It is associated with
${\boldsymbol{p}_X}^a$ and ${\mathbf{R}_X}^a$.
\item[${\mathbf{g}_X}^{p'}$] The force-closure grasps of object \textit{X} at
the pre-assembly positions.
\item[${\mathbf{g}_X}^{p}$] The colllision-free and IK
feasible grasps of object \textit{X} at the pre-assembly positions.
\end{itemize}

\section{3D Detection during Human Teaching}

\begin{figure}[!htbp]
  \centering
  \includegraphics[width = 3.4in]{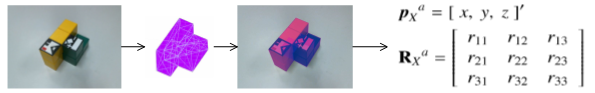}
  \caption{Object detection using AR markers. Beforehand, human worker
  attaches the markers and presaves the pose of the marker in the object's local
  coordinate system. During detection, vision system computes the pose of the
  object from the detected marker poses (the three subfigures). The output is
  the (${\boldsymbol{p}_A}^a$, ${\mathbf{R}_A}^a$) and (${\boldsymbol{p}_B}^a$,
  ${\mathbf{R}_B}^a$).}
  \label{ardetect}
\end{figure}

The object detection during human teaching is done using AR markers and a RGB
camera. Fig.\ref{ardetect} shows the flow of the detection and the poses of
the markers in the object model's local coordinate system. The markers are
attached manually by human workers.

During demonstration, the worker holds the two objects and show the makers to
the camera. We assume the workers have enough intelligence and can expose the
markers to the vision system without occlusion.
The detection process is conventional and can be found in many AR literature:
Given the positions of some features in the markers' local coordinate system,
find the transform matrix which converts the them into the positions on the
camera screen. In our implementation, the human teaching part is developed using
Unity and the AR recognition is done using the Vuforia SDK for Unity.

In the example shown in Fig.\ref{ardetect}, there are two objects where the
detected results are represented by (${\boldsymbol{p}_A}^a$,
${\mathbf{R}_A}^a$) and (${\boldsymbol{p}_B}^a$,
${\mathbf{R}_B}^a$). During robot execution, the (${\boldsymbol{p}_B}^a$,
${\mathbf{R}_B}^a$) is set to:
\begin{gather}
    {\boldsymbol{p}_B}^a
    =
    {\boldsymbol{p}_B}^a - {\boldsymbol{p}_A}^a, ~
    {\mathbf{R}_B}^{p}
    =
    {\mathbf{R}_B}^{a} \cdot ({\mathbf{R}_A}^{a})'
\end{gather}
and ${\boldsymbol{p}_A}^a$ and ${\mathbf{R}_A}^a$ are set to zero and identity
matrix respectively. Only the relative poses between the assembly parts are
used.

\section{3D Detection during Robotic Execution}

The object pose detection during robot execution is done using Kinect, Point
Cloud Library, and geometric constraints. The detection cannot be done using
markers since: (1) What the robot manipulates are thousands of industrial parts,
it is impossible to attach markers to all of them. (2) The initial configuration
of the object is unexpectable and the markers might be occluded from time to
time during robotic pick-and-place.
The detection is also not applicable to image-based approaches: Industrial
parts are usually mono colored and textureless, image features are not only
helpless but even harmful. 

\begin{figure}[!htbp]
  \centering
  \includegraphics[width = 3.4in]{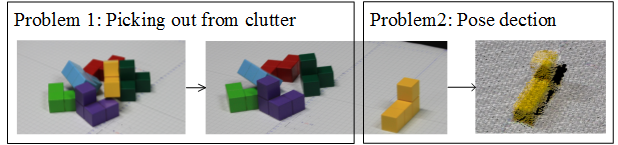}
  \caption{Overcome the clutter problem by dividing the pose detection into two
  subproblems. The first one is picking out from clutter where the system
  doesn't care what the object is or the pose of the object and its only goal is
  to pick something out. The problem is well studied. The second one is to
  detect the pose of a single object on an open plenary surface. It is not
  trivial since the precision of Kinect is bad.}
  \label{twoprb}
\end{figure}

Using Kinect is not trivial due to its low resolution and precision. For
example, the objects in industrial applications are usually in clutter and it is
difficult to segment one object from another using Kinect's resolution. Our
solution is to divide the difficulty in clutter into two subproblems. First, the
robot considers how to pick out one object from the clutter, and place the
object on an open plenary surface. Second, the robot estimate the pose of the
single object on the open plenary surface. The first subproblem doesn't care
what the object is or the pose of the object and its only goal is to pick
something out.
It is referred as a pick-and-place problem in contemporary literature and is
illustrated in the left part of Fig.\ref{twoprb}. The first subproblem is well
studied and interested readers may refer to \cite{Domae14} for the solutions.

The second subproblem is to detect the pose of a single object on an
open plenary surface and is shown in the right part of Fig.\ref{twoprb}. It is
much easier but still requires much proess to meet the precision requirements of
assembly. We concentrate on precision and will discuss use point cloud
algorithms and geometric constraints to solve the second problem.

\subsection{Rough detection using point cloud}

\begin{figure}[!htbp]
  \centering
  \includegraphics[width = 3.4in]{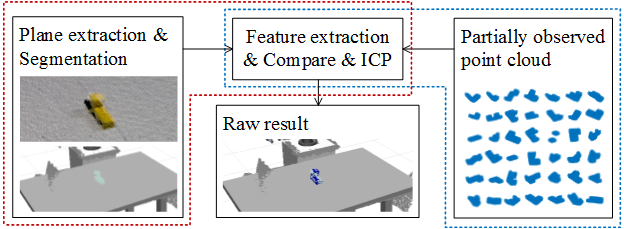}
  \caption{Rawly detecting the pose of an object using model matching and CVFH
  and CRH features. In a preprocessing process before the detection, we
  precompute the CVFH and CRH features of 42 different views and save them as
  the template. The preprocessing process is shown in the
  dashed blue frame. During the detection, we segment the remaining points
  cloud, compute the CVFH and CRH features of each segmentation, and match them
  to the precomputed views using ICP. The best match is used as the output.}
  \label{rawpose}
\end{figure}

First, we roughly detect the pose of the object using CVFH and CRH features.
In a preprocessing process before starting the detection, we put the camera to
42 positions on a unit sphere, save the view of the camera, and
precompute the CVFH and CRH features of each view. This step is virtually
performed using PCL and is shown in the dashed blue frame of Fig.\ref{rawpose}.
During the detection, we extract the plenary surface from the point
cloud, segment the remaining points cloud, and compute the CVFH and CRH features of
each segmentation. Then, we match the precomputed features with the features of
each segment and estimate the orientation of the segmentations. This step is
shown in the dashed red frame of Fig.\ref{rawpose}. The matched segments are
further refined using ICP to ensure good matching.
The segmentation that has highest ICP matches and smallest outlier points will
be used as the output. An example is shown in the ``Raw result''
framebox of Fig.\ref{rawpose}.

\subsection{Noise correction using geomteric constraints}

\begin{figure}[!htbp]
  \centering
  \includegraphics[width = 3.4in]{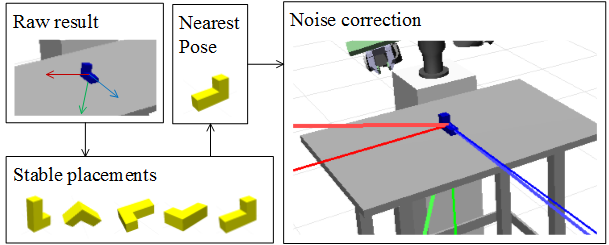}
  \caption{Correcting the raw result using the stable placements on
  a plenary surface (geometric constraints). In a preprocessing process before
  the correction, we compute the stable placements of the object on a plenary
  surface. An example is shown in the stable placements framebox. During the
  correction, we compute the distance between the raw results and each of the
  stable placements, and correct the raw results using the nearest pose.}
  \label{posecor}
\end{figure}

The result of rough detection is further refined using the geometric
constraints.
Since the object is on plenary surface, its stable poses are limited
\cite{Wan2016ral} and can be used to correct the noises of the roughly estimated
result. The placement planning includes two steps. First, we compute the convex
hull of the object mesh and perform surface clustering on the convex hull. Each
cluster is one candidate standing surface where the object may be placed on.
Second, we check the stability of the objects standing on these candidate
surfaces. The unstable placements (the placement where the projection of center
of mass is outside the candidate surface or too near to its boundary) are
removed.
An example of the stable placements for one object is shown in the 
stable placements framebox of Fig.\ref{posecor}.

Given the raw detection result using CVFH and CRH features, the robot computes
its distance to the stable placements and correct it following
Alg.\ref{ncalg}.

\begin{algorithm}[htbp!]
\DontPrintSemicolon
    \KwData{Raw result: $\boldsymbol{p}_r$, $\mathbf{R}_r$;\\
            ~~~~~~~~~~~Stable placements: \{$\mathbf{R}_p(i),~i=1,2,\ldots$\}\\
            ~~~~~~~~~~~Table height: $h_t$}
    \KwResult{Corrected result: $\boldsymbol{p}_c$, $\mathbf{R}_c$}
    \SetKwFunction{size}{size}
    \SetKwFunction{rpyFromRot}{rpyFromRot}
    \SetKwFunction{rotFromRpy}{rotFromRpy}
    $d_{min}\leftarrow+\infty$\;
    $\mathbf{R}_{near}\leftarrow\mathbf{I}$\;
    \For{$i\leftarrow1$ \KwTo $\mathbf{R}_p$.\size{}} {
        $d_i\leftarrow||\log(\mathbf{R}_p(i){\mathbf{R}_p}{'})||$\;
        \If{$d_i<d_{min}$} {
            $d_{min} \leftarrow d_i$\;
            $\mathbf{R}_{near} \leftarrow \mathbf{R}_p(i)$\;
        }
    }
    $\mathbf{R}_{yaw} \leftarrow$ \rotFromRpy{0, 0,
    \rpyFromRot{$\mathbf{R}_{r}$}$.y$}\;
    $\mathbf{R}_c\leftarrow\mathbf{R}_{yaw}\cdot\mathbf{R}_{near}$\;
    $\boldsymbol{p}_c\leftarrow[\boldsymbol{p}_r.x, \boldsymbol{p}_r.y,
    h_t]'$
    \caption{Noise correction}
    \label{ncalg}
\end{algorithm}

In this pseudo code, $\mathbf{I}$ indicates a $3\times3$ identity matrix.
Functions $\mathtt{rotFromRpy}$ and $\mathtt{rpyFromRot}$ converts $roll$,
$pitch$, $yaw$ angles to rotation matrix and vice verse. The distance between
two rotation matrices is computed in line 4. The corrected result is updated at
lines 11 and 12.

\section{Grasp and Motion Planning}

After finding the poses of the parts on the plenary surface, what the robot does
next is to grasp the parts and assemble them. It includes two steps: A grasp
planning step and a motion planning step. 

\subsection{Grasp planning}
In the grasp planning step, we set the
object at free space and compute the force-closured and collision-free grasps.
Each grasp is represented using ${\boldsymbol{g}_X}^f$=$\{\boldsymbol{p}_0,
\boldsymbol{p}_1, \mathbf{R}\}$ where $\boldsymbol{p}_0$ and
$\boldsymbol{p}_1$ are the contact positions of the finger tips, $\mathbf{R}$
is the orientation of the palm. The whole set is represented by
${\mathbf{g}_X}^f$, which includes many ${\boldsymbol{g}_X}^f$. Namely,
${\mathbf{g}_X}^f$ = $\{{\boldsymbol{g}_X}^f\}$.

Given the pose of a part on the plenary surface, say ${\boldsymbol{p}_X}^s$
and ${\mathbf{R}_X}^s$, the IK-feasible and collision-free grasps that the robot
can use to pick up the object is computed following
\begin{equation}
    {\mathbf{g}_X}^{s}
    =
    \mathsf{IK}~(~{\mathbf{g}_X}^{s'}~)
    \cap 
    \mathsf{CD}~(~{\mathbf{g}_X}^{s'},~planery~surface~)
    \label{transsurf1}
\end{equation}
where
\begin{equation}
    {\mathbf{g}_X}^{s'}
    =
    {\mathbf{R}_X}^s\cdot{\mathbf{g}_X}^f+{\boldsymbol{p}_X}^s 
    \label{transsurf2}
\end{equation}

${\mathbf{R}_X}^s\cdot{\mathbf{g}_X}^f+{\boldsymbol{p}_X}^s$ transforms the
grasps at free space to the pose of the object. ${\mathbf{g}_X}^{s'}$ denotes
the transformed grasp set. $\mathsf{IK}()$ finds the IK-feasible
grasps from the input set. $\mathsf{CD}()$ checks the collision between the two
input elements and finds the collision-free grasps. ${\mathbf{g}_X}^{s}$
denotes the IK-feasible and collision-free grasps that the robot can use to pick
up the object.

Likewise, given the pose of object A in the assembled structure, say
${\boldsymbol{p}_A}^a$ and ${\mathbf{R}_A}^a$, the IK-feasible and
collision-free grasps that the robot can use to assemble it is computed
following
\begin{equation}
    {\mathbf{g}_A}^{a}
    =
    \mathsf{IK}~(~{\mathbf{g}_A}^{a'}~)
    \cap 
    \mathsf{CD}~(~{\mathbf{g}_A}^{a'},\mathsf{objB}(~{\boldsymbol{p}_B}^a,
    {\mathbf{R}_B}^a~)~)
\end{equation}
where
\begin{equation}
    {\mathbf{g}_A}^{a'} =
    {\mathbf{R}_A}^a\cdot{\mathbf{g}_A}^f+{\boldsymbol{p}_A}^a \\
\end{equation}

${\mathbf{R}_A}^a\cdot{\mathbf{g}_A}^f+{\boldsymbol{p}_A}^a$ transforms
the grasps at free space to the pose of the object in the assembled structure.
${\mathbf{g}_X}^{a'}$ denotes the transformed grasp set. $\mathsf{IK}()$ 
and $\mathsf{CD}()$ are the same as Eqn(\ref{transsurf1}).
$\mathsf{objB}(~{\boldsymbol{p}_B}^a, {\mathbf{R}_B}^a~)$ indicates the mesh
model of object B at pose ${\boldsymbol{p}_B}^a, {\mathbf{R}_B}^a$. ${\mathbf{g}_A}^{a}$
denotes the IK-feasible and collision-free grasps that the robot can use to
assemble the object.

\subsection{Motion planning}
In the motion planning step, we build a graph using the elements in
${\mathbf{g}_X}^{s}$ and ${\mathbf{g}_X}^{a}$, search the grasp to find
high-level keyframes, and perform Transition-based Rapidly-Exploring Random
Tree (Transition-RRT) \cite{Jaillet08} motion planning between the keyframes to
find assemble motions.

\begin{figure}[!htbp]
  \centering
  \includegraphics[width = 3.4in]{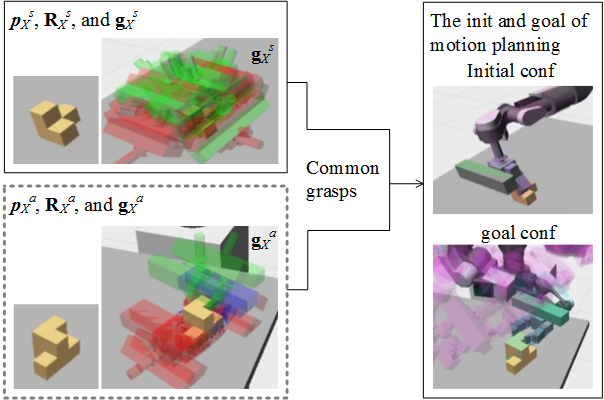}
  \caption{The flow of the motion planning. Given initial and goal poses of an
  object (left images in the upperleft and lowerleft frameboxes), we search its
  available initial and goal grasps and use the common grasps and IK to get the
  initial and goal configurations of the robot arm.
  Then, we do motion planning repeatedly between the initial and goal
  configurations to find a solution to the desired task.
  }
  \label{mflow}
\end{figure}

Fig.\ref{mflow} shows the flow. The object \textit{X} in this graph is a wooden
block shown in the left image of the upper-left frame box. The image also shows
the pose of this object on the plenary surface, ${\boldsymbol{p}_X}^s$ and
${\mathbf{R}_X}^s$. When the object is assembled in the structure, its pose
${\boldsymbol{p}_X}^a$ and ${\mathbf{R}_S}^a$ is shown in the left image of the
bottom-left frame box.
The grasps associated with the poses are shown in the right images of the two
frame boxes. They are rendered using the colored hand model. Green,
blue, and red denote IK-feasible and collision free,
IK-infeasible, and collided grasps respectively.
We build a graph to find the common grasps and employ Transition-RRT to find the
motion between the initial configuration and goal configuration.

In practice, the flow in Fig.\ref{mflow} doesn't work since the goal
configuration is in the assembled structure and is in the narrow passages or on
the boundaries of the configuration space. The motion planning problem is a
narrow-passage \cite{Wan2008} or peg-in-hole problem \cite{Yun2008} which is not
solvable. We overcome the difficulty by adding a pre-assemble configuration:
For the two objects A and B, we retract them from the structure following the
approaching direction $\boldsymbol{v}^a$ of the two objects and get the
pre-assemble poses ${\boldsymbol{p}_A}^p$, ${\mathbf{R}_A}^p$, and
${\boldsymbol{p}_B}^p$, ${\mathbf{R}_B}^p$ where
\begin{gather}
    {\boldsymbol{p}_A}^p
    =
    {\boldsymbol{p}_A}^a+0.5\boldsymbol{v}^a,~
    {\mathbf{R}_A}^{p}
    =
    {\mathbf{R}_A}^{a}
    \label{ret1}\\
    {\boldsymbol{p}_B}^p
    =
    {\boldsymbol{p}_B}^a-0.5\boldsymbol{v}^a,~
    {\mathbf{R}_B}^{p}
    =
    {\mathbf{R}_B}^{a}
    \label{ret2}
\end{gather}

The grasps associated with the retracted poses are

\begin{eqnarray}
    {\mathbf{g}_A}^p
    =
    \mathsf{IK}~(~{\mathbf{g}_A}^{p'}~),~\mathrm{where}~
    {\mathbf{g}_A}^{p'}
    =
    {\mathbf{g}_A}^a+0.5\boldsymbol{v}^a
    \label{retg1}\\
    {\mathbf{g}_B}^p
    =
    \mathsf{IK}~(~{\mathbf{g}_B}^{p'}~),~\mathrm{where}~
    {\mathbf{g}_B}^{p'}
    =
    {\mathbf{g}_B}^a-0.5\boldsymbol{v}^a
    \label{retg2}
\end{eqnarray}

Note that the poses in Eqn(\ref{ret1}-\ref{retg2}) are in the local coordinate
of object A where ${\boldsymbol{p}_A}^a$ is a zero vector and
${\boldsymbol{R}_A}^a$ is an identity matrix. Given the pose of object A in
world coordinate, ${\boldsymbol{p}_A}^{a(g)}$ and ${\boldsymbol{R}_A}^{a(g)}$,
the grasps in the world coordinate are computed using

\begin{gather}
    {\mathbf{g}_A}^{a(g)}={\mathbf{g}_A}^{p(g)}\\
    {\mathbf{g}_B}^{a(g)}=
    {\boldsymbol{p}_A}^{a(g)}+{\boldsymbol{R}_A}^{a(g)}\cdot{\mathbf{g}_B}^a\\
    {\mathbf{g}_A}^{p(g)}=
    {\boldsymbol{p}_A}^{a(g)}+{\boldsymbol{R}_A}^{a(g)}\cdot{\mathbf{g}_A}^p\\
    {\mathbf{g}_B}^{p(g)}=
    {\boldsymbol{p}_A}^{a(g)}+{\boldsymbol{R}_A}^{a(g)}\cdot{\mathbf{g}_B}^p
\end{gather}

The moton planning is then to find a motion between one initial configuration in 
${\mathbf{g}_X}^s$ to a goal configuration in ${\mathbf{g}_X}^p(g)$ where $X$
is either $A$ or $B$. There is no motion between ${\mathbf{g}_A}^p(g)$ and
${\mathbf{g}_A}^a(g)$ since they are equal to each other. The motion between
${\mathbf{g}_B}^p(g)$ and ${\mathbf{g}_B}^a(g)$ is hard coded along
$\boldsymbol{v}^a$.

\begin{figure}[!htbp]
  \centering
  \includegraphics[width = 3.2in]{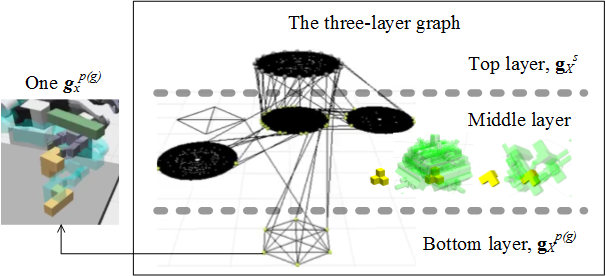}
  \caption{The grasp graph built using ${\mathbf{g}_X}^{s}$ and
  ${\mathbf{g}_X}^{p(g)}$. It has three layers where the top layer encodes the
  grasps associated with the initial configuration, the middle layer encodes
  the grasps associated with placements on planery sufaces, and the bottom
  layer encodes the grasps associated with the assemble pose. The left image
  shows one ${\mathbf{g}_X}^p(g)$ (the virtual grasp illustrated in cyan).
  It corresponds to a node in the bottom layer. The subimages in the frame box
  illustrate the placements (yellow) and their associated grasps (green).}
  \label{graph}
\end{figure}

Which initial and goal configuration to use is decided by building and searching
a grasp graph which is built using ${\mathbf{g}_X}^{s}$ and
${\mathbf{g}_X}^{p(g)}$, and is shown in the frame box of Fig.\ref{graph}. The
graph is basically the same as \cite{Wan2016ral}, but has three layers. The top
layer has only one circle and is mapped to the initial configuration. The bottom
layer also has only one circle and is mapped to the goal configuration. The
middle layers are composed of several circles where each of them maps a
possible placement on a plenary surface.
Each node of the circles represents a grasp: The ones in the upper layers are
from ${\mathbf{g}_X}^s$, and the ones in the bottom layers are from
${\mathbf{g}_X}^{p(g)})$. The ones from the middle layers are the grasps
associated with the placements. The orientations of the placements are evenly
sampled on line. Their positions are fixed to the initial position
${\mathbf{p}_A}^s$.
If the circles share some grasps (grasps with the same $\boldsymbol{p}_0$,
$\boldsymbol{p}_1$, $\mathbf{R}$ values in the object's local coordinate
system), we connect them at the correspondent nodes. We search the graph to find
the initial and goal configurations and a sequence of high-level keyframes, and
perform motion planning between the keyframes to perform desired tasks. An
exemplary result will be shown in the experiment section.

\section{Experiments and Analysis}

We performed experiments to examine the precision of the developed approaches,
analyze the process of grasp and motion planning,
and demonstrate the applicability of the study using a Kawada Nextage Robot.
The camera used in the human teaching phase is a logicool HD Webcam C615. The
computer system is a Lenovo Thinkpad E550 laptop (Processor: Intel Core i5-5200
2.20GHz Clock, Memory: 4G 1600MHz DDR3).
The depth sensor used in the robotic execution phase is Kinect. The computer
system used to compute the grasps and motions is a Dell T7910 workstation
(Processor: Intel Xeon E5-2630 v3 with 8CHT, 20MB Cache, and 2.4GHz Clock,
Memory: 32G 2133MHz DDR4).

\subsection{Precision of the object detection in human teaching}

First, we examine the precision of object detection during human teaching.
We use two sets of assembly parts and examine the precision of five assembly
structures for each set. Moreover, for each assembly structure, we examine the
values of at five different orientations to make the results confident.

\begin{figure*}[!htbp]
  \centering
  \includegraphics[width = 7.2in]{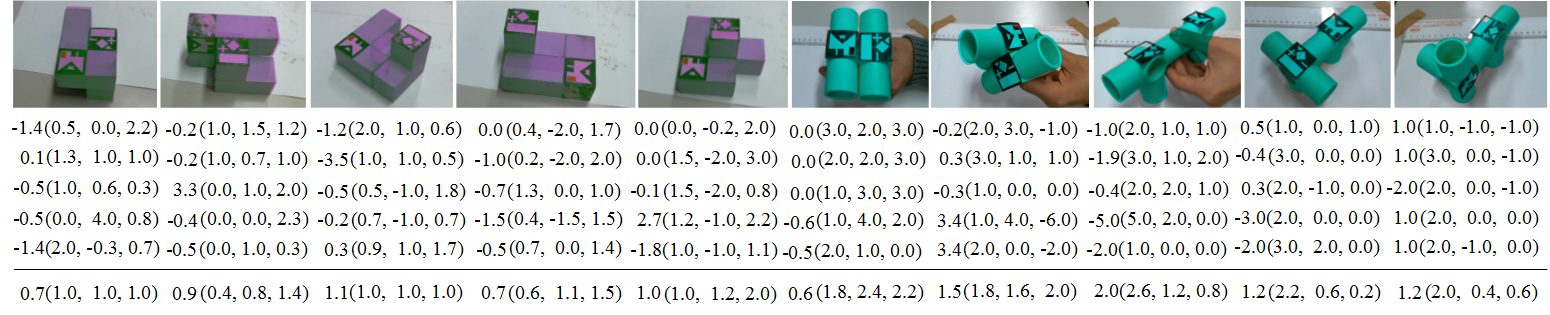}
  \caption{Results of the object detected during human teaching. The image
  row shows the structure to be assembled. Each structure is posed at five
  different orientations to examine the precision and the detected error
  in distance and orientation are shown in the five data rows
  below. Each grid of the data rows is shown in the form $\Delta d (\Delta r,
  \Delta p, \Delta y)$ where $\Delta d$ is the difference in the measured
  $|{\boldsymbol{p}_A}^{a}-{\boldsymbol{p}_B}^{a}|$ and the actual value.
  $(\Delta r, \Delta p, \Delta y)$ are the difference in the measured roll,
  pitch, and yaw angles. The last row is the average detection error. The
  metrics are $millimeter$ for distance and $degree$ for orientation.}
  \label{exphumandemo}
\end{figure*}

The two sets and ten structures (five for each) are shown in the first row of
Fig.\ref{exphumandemo}. Each structure is posed at five different orientations
to examine the precision. The five data rows under the subimage row in
Fig.\ref{exphumandemo} are the result at the orientations. Each grid of the
data rows is shown in the form $\Delta d (\Delta r, \Delta p, \Delta y)$ where
$\Delta d$ is the difference in the measured
$|{\boldsymbol{p}_A}^{a}-{\boldsymbol{p}_B}^{a}|$ and the actual value. $(\Delta
r, \Delta p, \Delta y)$ are the difference in the measured $roll$, $pitch$, and
$yaw$ angles. The last row of the figure shows the average detection error of
each structure in the form $|\Delta d| (|\Delta r|, |\Delta p|, |\Delta y|)$
where $|\cdot|$ indicates the absolute value. The metrics are $millimeter$
($mm$) for distance and $degree$ ($^\circ$) for orientation. The precision in
position is less than 1$mm$ and the precision in orientation is less than
2$^\circ$ on average.

\subsection{Precision of the object detection in robotic execution}

\begin{figure*}[!htbp]
  \centering
  \includegraphics[width = 7.2in]{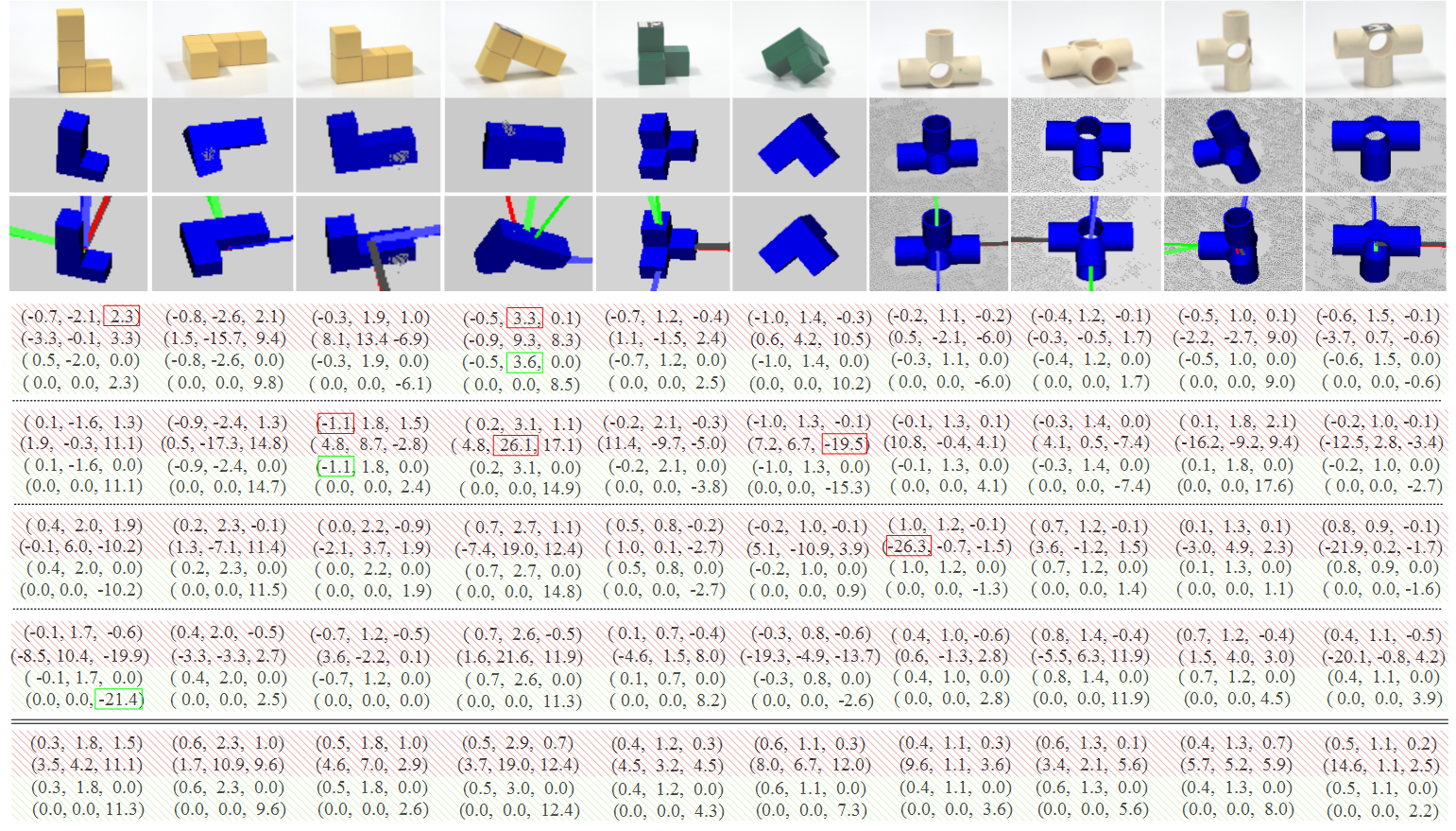}
  \caption{Results of the object detected during robotic execution. The figure
  includes three subfigure rows and five data rows where the first data row
  show the target object pose, the second and third subfigure rows plot some
  examples of the roughly detected poses and the corrected poses. The five data
  rows are divided by dashed or solid lines where the first four of them show
  the individual detection precision at four different positions and the last
  one shows the average detection precision. Each data grid of the data rows
  include four triples where the upper two (under red shadow) are the roughly
  detected position and orientation and the lower two (under green shadow) are
  the corrected values. The three values of the position triples are the $x$,
  $y$, $z$ coordinates, and their metrics are $millimeters$ ($mm$). The three
  values of the orientation triples are the $roll$, $pitch$, $yaw$ angles and
  their metrics are $degree$ ($^\circ$). The maximum values of each data
  element are marked with colored frameboxes.}
  \label{exprobotexe}
\end{figure*}

Then, we examine the pecision of the object detection in the robot execution
phase. Three objects with eight placements are used during the process. They are
shown in the top row of Fig.\ref{exprobotexe}. The plenary surface is set in
front of the robot and is divided into four quarters. We place each placement
into each quoter to get the average values. There are five
data rows divided by dashed or solid lines in Fig.\ref{exprobotexe} where the
first four of them show the individual detection precision at each quoter and
the last one shows the average detection precision. The detection precision is
the difference between the detected value and the groundtruth. Since we know the
exact model of the object and the height of the table, the groundtruth is know
beforehand. The average detection precision in the last row are the
mean of the absolute difference.

Inside each data grid there are four triples where the upper two are the
roughly detected position and orientation and the lower two are the corrected
values. The roughly detected results are marked with red shadows and the
corrected results are marked in green. The three values of the position
triples are the $x$, $y$, $z$ coordinates, and their metrics are
$millimeters$ ($mm$). The three values of the orientation triples are the
$roll$, $pitch$, $yaw$ angles and their metrics are $degree$ ($^\circ$).

The results show that the maximum errors of rough position detection are
-1.1$mm$, 3.3$mm$, and 2.3$mm$ along $x$, $y$, and $z$ axis. They are marked
with red boxes in Fig.\ref{exprobotexe}. After correction, the maximum position
errors change to -1.1$mm$, 3.6$mm$, 0.0$mm$ respectively. They are marked with
green boxes. The maximum errors of rough orientation detection are
-26.3$^\circ$, 26.1$^\circ$, and -19.5$^\circ$ in $roll$, $pitch$, and $yaw$
angles. They are marked with red boxes. The errors change to $0.0$, $0.0$, and
-$21.4$ after correction. The correction using geometric constraints completely
corrects the erros along $z$ axis and in $roll$ and $pitch$ angles. It might
slightly increase the errors along $x$ and $y$ and in $yaw$ but the values are
ignorable. The average performance can be found from the data under thee dual
solid line. The performance is good enough for robotic assembly.

In addition, the second and third subfigure rows of Fig.\ref{exprobotexe} plot
some examples of the roughly detected poses and the corrected poses.
Readers may compare them with the data rows.

\subsection{Simulation and real-world results}

\begin{figure*}[!htbp]
  \centering
  \includegraphics[width = 7.05in]{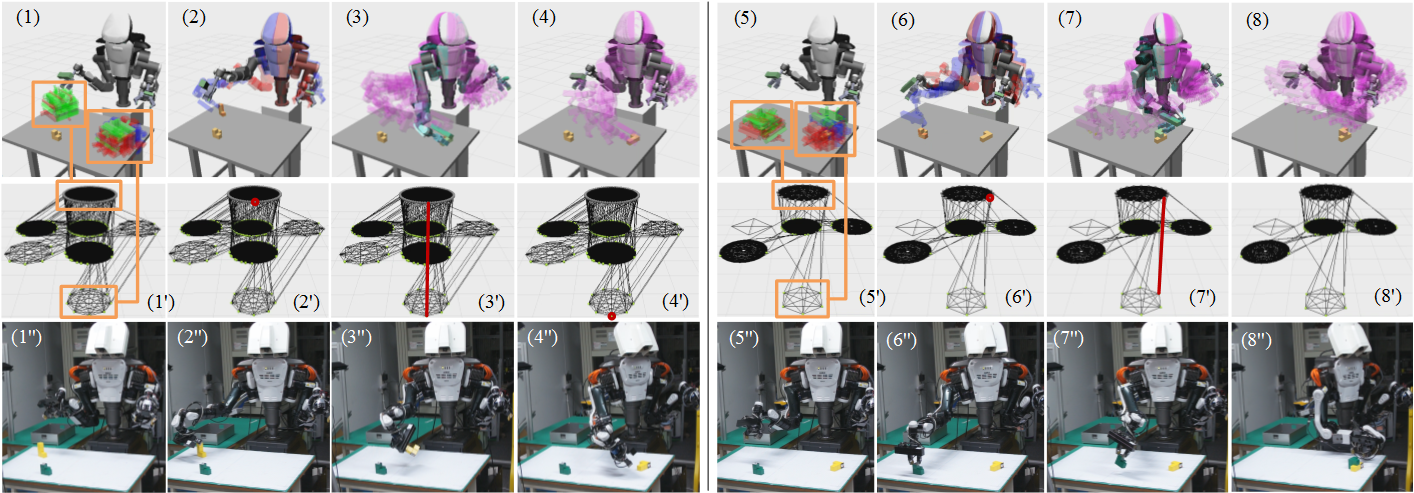}
  \caption{The snapshots of assembling the structure shown in
   Fig.\ref{exphumandemo}. It is divided into two step with the first step shown
   in (1)-(4) and the second step shown in (5)-(8). In the first step, the robot
   picks up object A and transfers it to the goal pose. In the second step,
   the robot finds object B on the table and assembles it to object A.
   The subfigures (1')-(8') shows correspondent path edges and nodes on the
   three-layer graph.}
  \label{expcorr}
\end{figure*}

Fig.\ref{expcorr} shows the correspondence between the paths found by the
searching algorithms and the configurations of the robot and the object. The
structure to be assembled in this task is the first one (upper-left one) shown in
Fig.\ref{exphumandemo}. We do motion planning along the paths repeatedly to
perform the desired tasks.


The assembly process is divided into two step with each step corresponds to one
assembly part. In the first step, the robot finds object A on the table and
moves it to a goal pose using the three-layer graph. The
subfigures (1)-(4) of Fig.\ref{expcorr} shows this step. In
Fig.\ref{expcorr}(1), the robot computes the grasps associated with the initial
pose and goal pose of object A. The associated grasps are rendered in green,
blue, and red colors like Fig.\ref{flow}. They corresponds to the top and bottom
layer of the grasp shown in Fig.\ref{expcorr}(1'). In
Fig.\ref{expcorr}(2), the robot chooses one feasible (IK-feasible and
collision-free) grasp from the associated grasps and does motion planning to
pick up the object. The selected grasp corresponds to one node in the top layer
of the graph, which is marked with red color in Fig.\ref{expcorr}(2'). In
Fig.\ref{expcorr}(3), the robot picks up object A and transfers it to the goal
pose using a second motion planning. This corresponds to an edge in
Fig.\ref{expcorr}(3') which connects the node in one circle to the node in
another. The edge directly connects to the goal in this example and there is no
intermediate placements. After that, the robot moves its arm back at
Fig.\ref{expcorr}(4), which corresponds to a node in the bottom layer of the
graph shown in Fig.\ref{expcorr}(4').

In the second step, the robot finds object B on the table and
assembles it to object A. The subfigures (5)-(8) Fig.\ref{expcorr}(b) show it.
In Fig.\ref{expcorr}(5), the robot computes the grasps associated with the
initial pose and goal pose of object B. They are rendered in
green, blue, and red colors like Fig.\ref{expcorr}(1) and are correspondent
to the top and bottom layer of the grasp shown in Fig.\ref{expcorr}(5'). In
Fig.\ref{expcorr}(6), the robot chooses one feasible grasp and does motion
planning to pick up the object. The selected grasp corresponds to the marked
node in Fig.\ref{expcorr}(6'). In Fig.\ref{expcorr}(7), the robot picks
up object B and assembles it to the goal pose using a second motion planning
which corresponds to an edge in Fig.\ref{expcorr}(7'). Finally, the robot
moves its arm back at Fig.\ref{expcorr}(8) and (8').

The subfigures (1'')-(8'') in the third row show how the robot executes the
planned motion. They correspond to (1)-(8) and (1')-(8') in the first two rows.

\section{Conclusions and Future Work}

We presented precise 3D visual detection approaches in this paper to meet the
requirements of a smart mechanical assembly system. In the human teaching phase
of the system where human beings control the operation and can actively avoid
occlusion, we use AR markers and compute the pose of the object by detecting
the markers' poses. In the robot execution phase where occlusions happen
unexpectedly, we use point cloud matching to find a raw pose and use extrinsic
constraints to correct the noises. We examine the precision of the approaches
in the experiment part and demonstrate that the precision fulfills assembly
tasks using a graph model and an industrial robot.

The future work will be on the manipulation and assembly aspect. The current
result is position-based assembly. It will be extended to force-based assembly
tasks like inserting, snapping, etc., in the future.

\bibliographystyle{elsarticle-num}
\bibliography{references}






\end{document}